\documentclass{article}

\usepackage[preprint]{corl_2024}
\usepackage{amsmath}
\usepackage{algorithm}
\usepackage{algorithmicx}
\usepackage[]{amsfonts}
\usepackage{algpseudocode}
\usepackage{multirow}
\usepackage{makecell}
\usepackage{wrapfig}
\usepackage{graphicx}
\usepackage{caption}
\usepackage{enumitem}
\usepackage{booktabs}
\usepackage{multirow}
\usepackage{mathtools}
\usepackage{cleveref}
\usepackage{tikz}

\usepackage{subfigure}

\DeclarePairedDelimiter{\norm}{\lVert}{\rVert}

\newcommand{\ImObs}{x}
\newcommand{\dataset}{\mathcal{D}}
\newcommand{\ProObs}{\theta}
\newcommand{\EmbImObs}{z_x}
\newcommand{\EmbProObs}{z_\theta}
\newcommand{\lan}{l} %
\newcommand{\lanleqt}{l} %

\title{LIMT: Language-Informed Multi-Task\\
Visual World Models}

\newcounter{inlineenum}
\renewcommand{\theinlineenum}{\alph{inlineenum}}
\newenvironment{inlineenum}
  {\unskip\ignorespaces\setcounter{inlineenum}{0}%
   \renewcommand{\item}{\refstepcounter{inlineenum}{\textit{\theinlineenum})~}}}
  {\ignorespacesafterend}

\author{Elie Aljalbout$^{1,2, *}$, Nikolaos Sotirakis$^{1,3, *}$, Patrick van der Smagt$^{1,4}$, Maximilian Karl$^{1}$, Nutan Chen$^{1}$
}

\begin{document}
\maketitle
\footnotetext{$^1$During this work, all authors were affiliated with the Machine Learning Research Lab at Volkswagen Group, Germany. $^2$E.A. is currently with the Robotics and Perception Group, at the Department of Informatics of the University of Zurich (UZH) and the Department of Neuroinformatics at UZH and ETH Zurich, Switzerland.
$^3$Technical University of Munich, Germany.
$^4$Faculty of Informatics, E\"otv\"os Lor\'and University, Budapest, Hungary.
$^*$Shared first authorship.}

\begin{abstract}
    Most recent successes in robot reinforcement learning involve learning a specialized single-task agent.
    However, robots capable of performing multiple tasks can be much more valuable in real-world applications.
    Multi-task reinforcement learning can be very challenging due to the increased sample complexity and the potentially conflicting task objectives.
    Previous work on this topic is dominated by model-free approaches.
    The latter can be very sample inefficient even when learning specialized single-task agents.
    In this work, we focus on model-based multi-task reinforcement learning.
    We propose a method for learning multi-task visual world models, leveraging pre-trained language models to extract semantically meaningful task representations.
    These representations are used by the world model and policy to reason about task similarity in dynamics and behavior.
    Our results highlight the benefits of using language-driven task representations for world models and a clear advantage of model-based multi-task learning over the more common model-free paradigm.
    
\end{abstract}

\keywords{Multi-Task Learning, Language-Conditioned World Models, Model-Based Reinforcement Learning}

\section{Introduction}

Reinforcement learning~(RL) methods have shown great potential in various robotic control tasks such as manipulation and locomotion~\citep{akkaya2019solving, hwangbo2019learning}.
The majority of successes in this domain are in single-task settings, where the agent is concerned with finding control policies for a single task.
Ideally, a single agent should be capable of performing various tasks and smoothly switching between different task performances.
This is especially important when considering the high cost of robotic systems such as manipulators.
The goal of multi-task reinforcement learning~(MTRL) is to learn such a single policy capable of performing multiple tasks by jointly optimizing the individual task objectives~\citep{brunskill2013sample}.
This joint training process can be beneficial, for instance, in terms of bootstrapping the learning of complex tasks~\citep{kalashnikov2022scaling}.
However, it can be highly challenging, not only due to the increased complexity of the problem, but also because different tasks can have conflicting objectives leading to unstable training. 
The majority of research on MTRL considers model-free methods.
However, model-free RL methods are considerably sample-inefficient even in single-task learning.
This property is undesirable in robot learning systems, where environment interactions are very expensive and hard to obtain in the real world.
Alternatively, it is possible to train robotic control policies in simulation and transfer them to the real world.
However, this process is not trivial and presents multiple challenges~\citep{pmlr-v78-james17a,akkaya2019solving, peng2018sim,aljalbout2024role}.
This efficiency problem becomes even more pronounced when dealing with high-dimensional and complex observations, such as the ones encountered in visual RL.
The extension to the more complex multi-task setting can further exacerbate these problems.

Model-based reinforcement learning~(MBRL) methods tend to have superior sample efficiency compared to the model-free approach~\citep{dulac2019challenges, dreamer}.
This boost in efficiency is achieved by incorporating a model of the environment, which allows the agent to simulate environment interactions for policy search. 
This capability is crucial for multi-task learning as it enables the agent to effectively share knowledge across tasks, reducing the overall amount of data needed to achieve proficient performance on each task.
By learning a shared model that captures the dynamics relevant to multiple tasks, the agent can develop a more holistic understanding of its embodiment and its environment dynamics. 

In this work, we propose a model-based vision-based method for multi-task learning.
Our method incorporates language-conditioning in the world model as well as the actor and critic.
These models are conditioned on language embeddings from a pre-trained language model.
By doing so, we leverage these semantically meaningful task representations to boost the parameter sharing in the world model and policy.
We evaluate the proposed approach using multiple robotic manipulation tasks from the CALVIN dataset~\citep{mees_calvin_2022}.
We compare our work to baselines from single-task MBRL, and model-free MTRL methods.
Our experiments demonstrate successful learning of a multi-task world model and its usage for learning a policy for multiple manipulation tasks.
We validate the benefit of using language-embeddings as task representations for the world model and the policy, and demonstrate a substantial performance improvement in comparison to model-free MTRL.

\begin{figure}
    \centering
    \includegraphics[width=\linewidth]{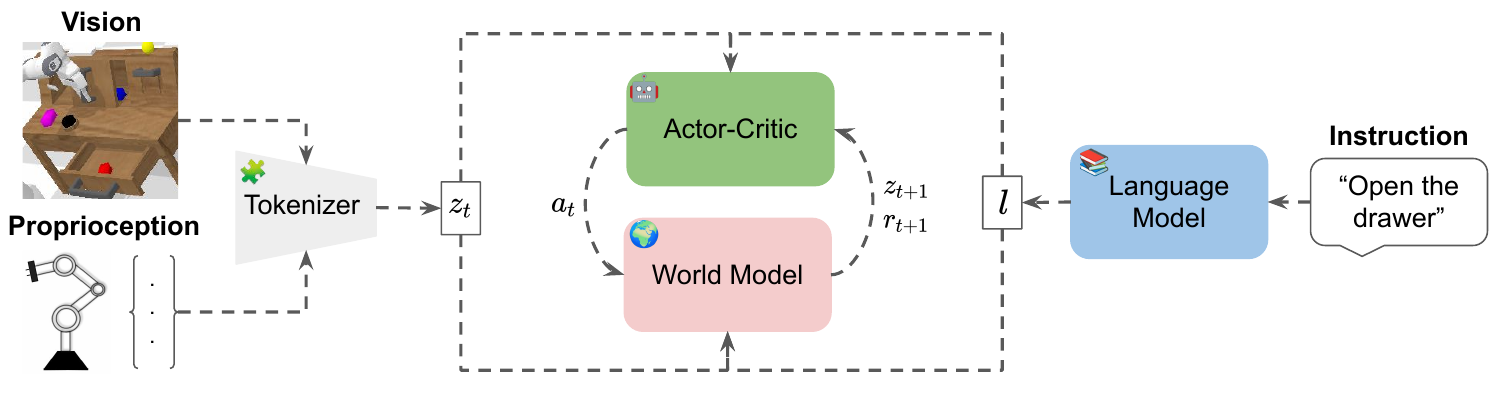}
    \caption{We train a model-based actor-critic agent using a multi-task world model. 
    The actor, critic and the world model are conditioned on tokens from a vision and proprioception tokenizer as well as a language embedding from a pretrained language model. 
    The latter encodes instructions for the different tasks.
    The resulting agent can perform multiple tasks depending on the instruction input.}
    \label{fig:overview}
\end{figure}

\section{Method}
Previous work has shown the benefits of multi-task policy training for bootstrapping the learning of harder tasks and increasing the individual tasks' sample efficiency~\citep{kalashnikov2022scaling}.
We postulate that a similar benefit can be observed for learning (visual) world models for robotics.
Intuitively, multiple tasks share similar dynamics and perception components.
For instance, a world model trained on opening a drawer has much in common with another trained on opening cupboards.
Training a common world model can leverage these task similarities to boost sample efficiency by sharing data across tasks.
To allow the model to reason about task similarity, a semantically meaningful task representation can be essential.
Hence, we propose a model-based MTRL approach, that trains a language-conditioned world model and actor-critic agent. 
Our approach consists of four main components. 
A \textbf{language model} encodes text instructions into structured embeddings.
A \textbf{tokenizer} maps observations to discrete token representations. 
A \textbf{world model} is a sequence model that predicts future observations, rewards, and successes based on past trajectories and task instruction embeddings. 
The \textbf{actor-critic} networks respectively output control commands and value estimates based on the latent state and embedding of the task description. 
We refer to our method as LIMT, an acronym for \textbf{l}anguage-\textbf{i}nformed \textbf{m}ulti-\textbf{t}ask visual world models.
Our overall approach is illustrated in \Cref{fig:overview}.

\subsection{Tokenizer}
Following the work in~\citep{iris}, we use a discrete autoencoder $(E, D)$ which is a variant of Vector Quantized Variational Autoencoder (VQVAE)~\citep{vqvae}, equipped with attention blocks as proposed in~\citep{vqgan} and additionally trained with a perceptual loss~\citep{vqgan}~\citep{perceptual_johnson}~\citep{larsen_learned_similarity}. The reason we chose to use discrete representations is that transformer networks, such as the one underlying our world model, are particularly successful at modeling sequences of discrete tokens~\citep{devlin2018bert}~\citep{gpt3}.
We further extend it to also handle proprioception data. 
The encoder $E$ accepts an observation $(\ImObs, \ProObs)$ consisting of an image observation $\ImObs$ and a $d$-dimensional proprioception vector $\ProObs$ and converts these to $K = K_\ImObs + K_\ProObs $ tokens of dimension $d_\mathrm{enc}$. Specifically, we use $K_\ImObs$ tokens to represent the image and $K_\ProObs$ tokens to represent proprioception. 
$E$ maintains two separate codebooks  of the same vocabulary size $N$ for the image and proprioception tokens respectively, $C_\ImObs =\{ c_\ImObs^i\}_{i=1}^N$, $C_\ProObs = \{c_\ProObs^i\}_{i=1}^N$, where $c_{\ImObs,\ProObs}^i \in \mathbb{R}^{d_\mathrm{enc}}$.

Concretely, the input of $E$ consists of observations $\ImObs \in \mathbb{R}^{H \times W \times C}$, $\ProObs \in \mathbb{R}^d$, where $H,W,C$ denote the image height, width, and number of channels respectively, and $d$ is the dimension of the proprioception vectors. 
$E$ passes $\ImObs$ through a series of convolutional and self-attention layers to obtain features $\EmbImObs \in \mathbb{R}^{h \times w \times d_\mathrm{enc}}$, where $h \times w = K_\ImObs$. 
It then computes a quantized embedding representation according to the nearest neighbor in $C_x$ using the Euclidean distance
\begin{equation}
    q_\ImObs\left((\EmbImObs)_{ij}\right) = \arg \min_{c\,\in\,C_\ImObs} {\left\vert (\EmbImObs)_{ij} - c\,\right\vert}^2 \quad ,\ i \in [h]\,,\ j \in [w].
\end{equation}

$E$ spatially decomposes the image into $h \times w$ feature vectors and assigns an element of the codebook to each one of them. 
Similarly, the proprioception vector $\ProObs$ is linearly projected to a latent vector $z_\ProObs \in \mathbb{R}^{d_\mathrm{enc}}$ via an affine layer, before being quantized according to the codebook $C_\ProObs$
\begin{equation}
    q_\ProObs(\EmbProObs) = \arg \min_{c\,\in\,C_\ProObs} {\left\vert \EmbProObs - c\,\right\vert}^2.
\end{equation}

The output $E(\ImObs,\ProObs) = (q_\ImObs(z_\ImObs), q_\ProObs(z_\ProObs))$ of the encoder comprises the latent representation of the observations, which is subsequently used as an input to the other components, namely the world model and the actor-critic. 
Since the set of possible encoder outputs is discrete, we can define the token representation of an observation $(\ImObs,\ProObs)$ as $w = (w_\ImObs, w_\ProObs)$, where $w_\ImObs, w_\ProObs \in \{1,..,N\}$.
For training purposes, a decoder $D$ maps these embeddings back into the observation space. 
For the RGB images, $D$ uses a network consisting of multiple convolutional, self-attention and upsampling layers, while the proprioception data is decoded via a single linear layer. 
We train our discrete autoencoder using the loss
\begin{align}
    L_A (E,D,\ImObs,\ProObs) = &\ \norm{ \ImObs - D(E(\ImObs)) }_1 + \norm{D(E(\ProObs)) - \ProObs}_2^2 \label{eq:lossrecon}\\
     +&\ \norm{ sg(E(\ImObs,\ProObs)) - z_{\ImObs,\ProObs} }_2^2 \label{eq:commitment}
    + \norm{ E(\ImObs,\ProObs) - sg(z_{\ImObs,\ProObs}) }_2^2 \\
    +&\ L_{\mathrm{perceptual}} (\ImObs, D(E(\ImObs))), \label{eq:percptual}
\end{align}

where $sg()$ is the stop-gradient operator. 
The right side of \eqref{eq:lossrecon} denotes the reconstruction loss for images and proprioception respectively, while the terms in~\eqref{eq:commitment} constitute the commitment loss, which ensures that the unquantized latent vectors are close to their corresponding discrete representations. 
\eqref{eq:percptual} is a perceptual loss~\citep{perceptual_johnson}~\citep{larsen_learned_similarity} shown in equation\eqref{eq:perceptual_loss} in the appendix, and computed with a pre-trained VGG16 CNN and given the ground-truth and the reconstructed images as inputs. 

\subsection{Language Model}
\label{sec:lm}

We use a pre-trained version of Sentence-BERT~\citep{reimers_sentence-bert_2019}, MiniLM-L6-v2 a large language model that has been tuned on semantic similarity, to encode natural language directives. In~\citep{mees_what_2022}, SBERT embeddings were found to be more suitable for language-conditioned policy learning compared to alternatives such as BERT~\citep{devlin2018bert} and CLIP~\citep{clip} embeddings.
We select this type of language model because it has a semantically meaningful structure of the embedding space, where encoded sentences can be compared using cosine similarity. 
Our hypothesis is that the embeddings of language instructions describing the same task or tasks with similar dynamics cluster nicely together and away of dissimilar tasks. 
This is shown in \Cref{fig:results}.
This natural clustering can help our agent reason about similarities among different tasks and thus learn skills and dynamics faster. 

\subsection{Dynamics Learning}
The world model, denoted as $G$, is an autoregressive transformer similar to the one in \citep{iris}.
In addition to actions and observation tokens, the transformer is conditioned on language embeddings. 
Given a trajectory of $T$ timesteps $(\ImObs_t, \ProObs_t, a_t, \lan')_{t=\tau}^T$, where
$\lan'$ denotes the language instruction, we first compute the sequence $(w_t, a_t, \lan)_{t=\tau}^T$, where $w^k_t \in \{1,..,N\}^{K}$ is the joint tokenized representation of the observations $\ImObs_t, \ProObs_t$, and $\lan$ is the instruction embedding from our language model. %
$G$ predicts the next observation tokens $\hat{w}_{t+1}$, the reward $\hat{r_t}$ and the episode end $\hat{d_t}$
\begin{gather}
     \hat{w}_{t+1} \sim p_G (\hat{w}_{t+1} \mid w_{\leq t}, a_{\leq t}, \lanleqt) \notag \\
    \hat{r_t} \sim p_G(\hat{r_t} \mid w_{\leq t}, a_{\leq t}, \lanleqt)  \\
    \hat{d_t} \sim p_G(\hat{d_t} \mid w_{\leq t}, a_{\leq t}, \lanleqt). \notag
\end{gather}

The transformer predicts the tokens $\hat{w}_{t+1}$ at $t+1$ autoregressively
\begin{equation} 
\label{eq:prediction}
    \hat{w}_{t+1}^{k+1} \sim p_G(\hat{w}_{t+1}^{k+1} \mid w_{t+1}^{\leq k}, w_{\leq t}, a_{\leq t}, \lanleqt).
\end{equation}

$G$ operates on a context window $H \in \mathbb{N}$, such that only the last $H$ timesteps influence the predictions. 
Besides the task-dependent reward and termination condition, the prediction of the next latent state is also conditioned on the language directive. %
The loss function for $G$ is
\begin{equation}
\label{eq:loss_world_model}
\begin{split}
        L_G = \sum_{t=\tau}^{\tau + H} \left[ \sum_{k=1}^K  w_{t+1}^{k+1} \log p_G (w_{t+1}^{k+1} \mid  w_{t+1}^{\leq k}, w_{\leq t}, a_{\leq t}, \lanleqt) \right] \\
        + \rho \norm{ r_t - \hat{r}_t }_2^2 + d_t \log p_G (\hat{d}_t \mid w_{\leq t}, a_{\leq t}, \lanleqt),
\end{split}
\end{equation}
where the first term on the right side of the equation is the cross entropy loss between predicted and ground-truth tokens. 
The second term is the reward prediction. 
The third is a cross-entropy loss for the termination label. 
$\rho \in \mathbb{R}$ is a hyperparameter for weighing reward loss against the other terms.

\subsection{Policy Learning}

The actor and critic networks receive the observation token embeddings as inputs, concatenated with the instruction embeddings, and the predicted proprioception information. 
They are jointly trained in latent imagination as in Dreamer~\citep{dreamer}. 
Given an encoded instruction $l$ and observation token embedding $q(z_{t_0})$ at timestep $t_0$, we start a rollout of length $H$: For each $0<t<H$, the actor outputs an action $a_t$, based on $l, w_t$ and the world model predicts the next observation embedding $q(z_{t+1})$ and reward $r_{t+1}$ which result from taking the action $a_t$. 
The critic is trained to regress the $V^\lambda$ estimates, defined in equation \eqref{eq:lambda_returns} in the appendix. %
To stabilize training, we maintain a target critic $\hat{v}_\psi$, the weights of which we periodically update with those of our value function $v_\psi$. $\hat{v}_\psi$ is used to compute the $V^\lambda$ returns, and the objective of our value function $v_\psi$ is to estimate them.

\subsection{Training Algorithm}
\label{sec:algorithm}
We begin our training using the offline episodes, $\dataset_\mathrm{offline}$ provided by CALVIN. In this work, we concentrate on a subset of tasks $\mathcal{T} = \{ T_i \}_i^N$. We gather episodes where one of the tasks in $\mathcal{T}$ is completed in a separate dataset $\dataset_{\mathrm{filtered}}$, used in the later stages of the training process.
First, we train our tokenizer on  the image-proprioception pairs of $\dataset_\mathrm{offline}$ for $N_t$ epochs until we observe convergence. 
Subsequently, we train our world model on $\dataset_\mathrm{offline}$ jointly with the tokenizer for another $N_w$ epochs.  
Given that $\dataset_\mathrm{offline}$ contains numerous episodes from tasks other than those in $\mathcal{T}$, we modify each of these episodes during the training of our world model. Specifically, we relabel every such episode with a task chosen from $\mathcal{T}$ with equal probability, adjusting the ground-truth rewards and language instructions accordingly. 
This increases the number of learning samples for our tasks of interest and provides negative examples, which are otherwise not available in the expert play data.
In the next stage, we train the tokenizer and world model for an additional $N_f$ epochs on $\dataset_{\mathrm{filtered}}$.
After that, we start performing rollouts using the actor's policy, which we collect in an online buffer $\dataset_\mathrm{online}$. During this phase, we jointly train the tokenizer, the world model and the actor-critic on both $\dataset_\mathrm{online}$ and $\dataset_{\mathrm{filtered}}$. We balance sampling from the two datasets using an adaptive ratio $p_\mathrm{online}$ denoting the proportion of samples from $\dataset_\mathrm{online}$. $p_\mathrm{online}$ starts at 0 and grows along with the size of $\dataset_\mathrm{online}$ until it reaches a maximum ratio $p_\mathrm{max}$.
Within each sampled episode, we select a starting state from which our agent imagines a trajectory uniformly at random across the time dimension.
We randomize the task goal by taking the episode's true goal with provability $p$ and randomly sampling another goal from $\mathcal{T}$ with probability $1-p$, using weights $w$.
In this context, specifying a goal task involves selecting a language instruction corresponding to the task as an input to our policy.

\section{Experimental Results}
\label{sec:result}

We design experiments to answer the following questions:
\begin{enumerate}[label=(Q\arabic*), ref=(Q\arabic*), leftmargin=*, itemsep=0.05em, nosep]
    \item Can we learn multi-task policies with language-informed world models and MBRL?
    \item Is MBRL more sample efficient than model-free RL in multi-task settings?
    \item Is multi-task training beneficial for MBRL for single and multi-task performance?
    \item Are language embeddings good task representations for learning multi-task world models?
 \end{enumerate}

\begin{figure}
    \centering
    \includegraphics[width=\linewidth]{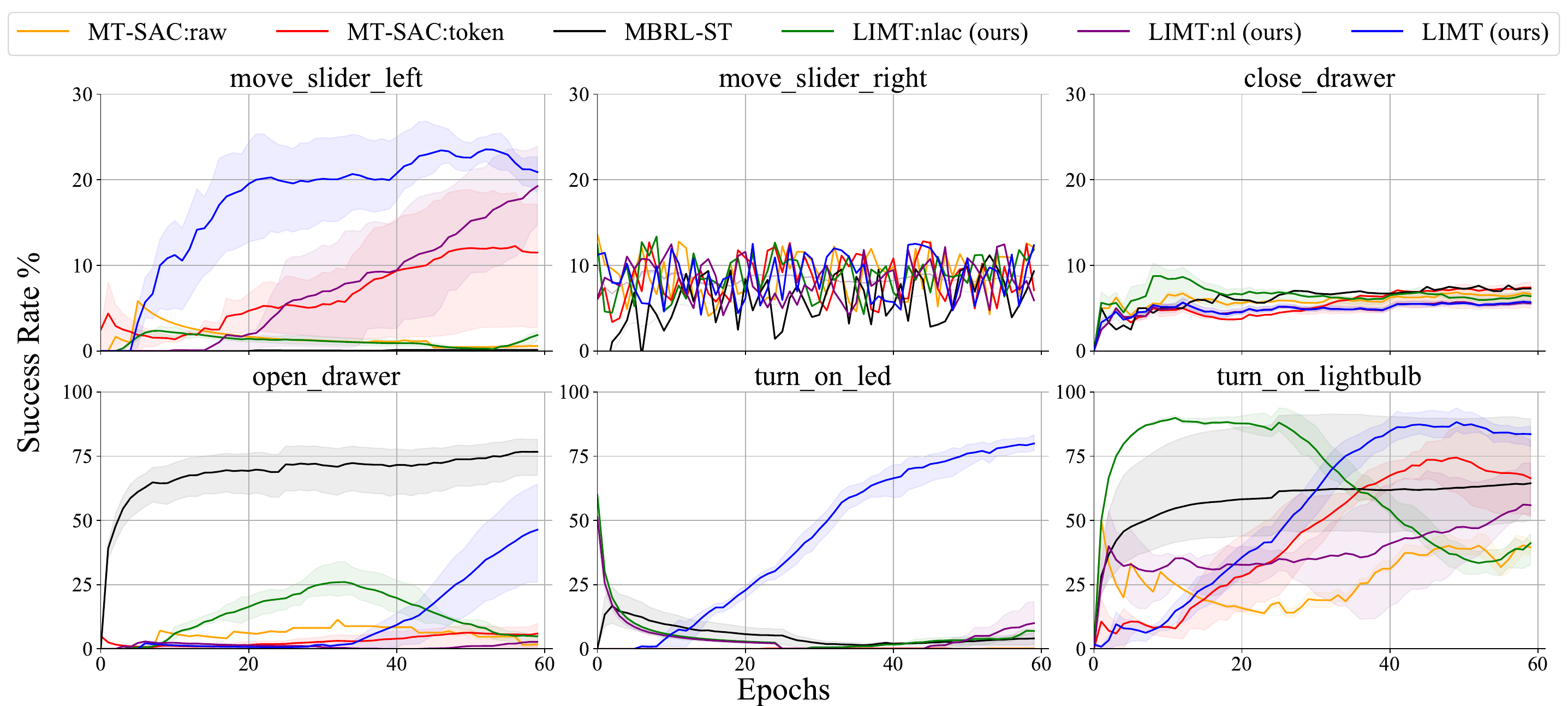}
    \caption{We plot the success rate over a fixed budget of training epochs. 
    We consider model-free baselines based on multi-task soft-actor-critic (MT-SAC).  
    MT-SAC is trained on raw images (MT-SAC:raw) and another variant is trained on image embeddings from our tokenizer (MT-SAC:token).
    We include a single-task MBRL baseline (MBRL-ST) and variants of our method that use integer task representations instead of language embeddings. The integer representations are either only used in the actor critic (LIMT:nlac) or in both the world model and actor-critic (LIMT:nl). 
    LIMT and its variant have better sample efficiency and success rate.}
    \label{fig:individual-tasks}
    \vspace{-0.3cm}    
\end{figure}

\subsection{Setup}

We train and deploy our agent on environment D of the CALVIN benchmark~\citep{mees_calvin_2022}.
The agent controls a 7-DoF Franka Emika Panda robot, equipped with a parallel gripper. It outputs actions in the form of relative Cartesian positions.
The environment consists of a table where the following static objects are present: i) a drawer that can be opened or closed, ii) a slider that can be moved left or right, iii) a button that toggles a green LED light iv) a switch that can be flipped up to control a lightbulb.
In addition to the static objects, 3 colored rectangular blocks appear in the scene in different positions and orientations. The agent's state consists of RGB images from a static camera, as well as 7-dimensional proprioception information indicating the position of the end effector and gripper width. We focus and evaluate our model on the following 6 tasks: \texttt{open\_drawer}, \texttt{close\_drawer}, \texttt{move\_slider\_left}, \texttt{move\_slider\_right}, \texttt{turn\_on\_lightbulb}, \texttt{turn\_on\_led}. 
For each of the 6 tasks, CALVIN~\cite{mees_calvin_2022} provides 7-14 distinct but semantically equivalent training language instructions. We leverage the embeddings of those instructions to train our agent. CALVIN also provides a single validation language directive for each task, which is distinct but equivalent to the training instructions. We use the validation directives when evaluating the performance of our agent.
During evaluation, we estimate our evaluation metrics, using 20 rollouts per task. 
We then append the rollouts to our online training dataset $\dataset_\mathrm{online}$ and continue training as described in \Cref{sec:algorithm}.
\subsection{Baselines}
\label{sec:baselines}
We compare our approach to  the following baselines:
\begin{inlineenum}
    \item MT-SAC:raw a model-free RL algorithm extending the SAC algorithm~\citep{sac} to multi-task settings.
    The policies are learned from raw images.
    To ensure a fair comparison, we use the same network architecture used in our method for all policy networks, including the actors, the critics and the targets.
    \item MT-SAC:token extends the raw version to use the latent states obtained from our tokenizer, as well as the language embeddings as inputs to the actor and critic. 
    \item MBRL-ST trains separate single-task (ST) world models and policies for each task. 
    \item LIMT:nlac is based on our method, but replaces the instruction embeddings with predefined integer task identifiers in the actor-critic networks. nlac refers to having no language in the actor critic.
    \item  LIMT:nl is based on our method, but replaces the instruction embeddings with predefined integer task identifiers in the world model as well as in the policy networks. nl refers to having no language in the whole model.
\end{inlineenum}
For a fair comparison, in the last two baselines, we repeat the task identifier multiple times to ensure it has the same dimension as the language embeddings.

\subsection{Results}
\label{sec:subresults}

\begin{figure}
    \centering
    \begin{minipage}{0.39\textwidth}
        \centering
        \includegraphics[width=\textwidth]{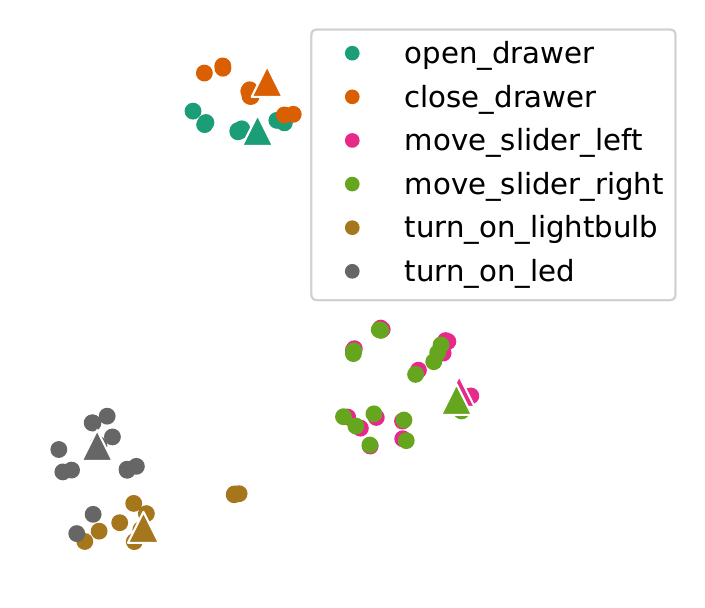} %
        \label{fig:example}
    \end{minipage}%
\begin{minipage}{0.57\textwidth}
    \centering
    \begin{tabular}{lc}
    \toprule
    \textbf{baseline} & \textbf{multi-task} \\ 
                      & \textbf{ success rate (\%)} \\
    \midrule
    MT-SAC:raw & 22.50 \\
    MT-SAC:token & 22.99 \\
    LIMT:nl & 24.01 \\
    LIMT:nlac & 27.15 \\
    LIMT (ours) & \textbf{52.12} \\ 
    \bottomrule
    \end{tabular}
    \label{tab:baseline}
\end{minipage}
\vspace{-0.7cm}
\caption{(Left) We visualize the language embedding of our different task instructions in 2-dimension using t-SNE \cite{van2008visualizing}. 
The circles and triangles represent instructions in the training and test datasets, respectively. 
Language embeddings of task instructions with semantic similarity tend to cluster well together.
(Right) We compare the multi-task success rate of our method to the studied baselines. 
LIMT outperforms all baselines in the multi-task setting by a large margin.}
\label{fig:results}
\vspace{-0.5cm}
\end{figure}

\begin{figure}
    \centering
    \includegraphics[width=0.95\linewidth]{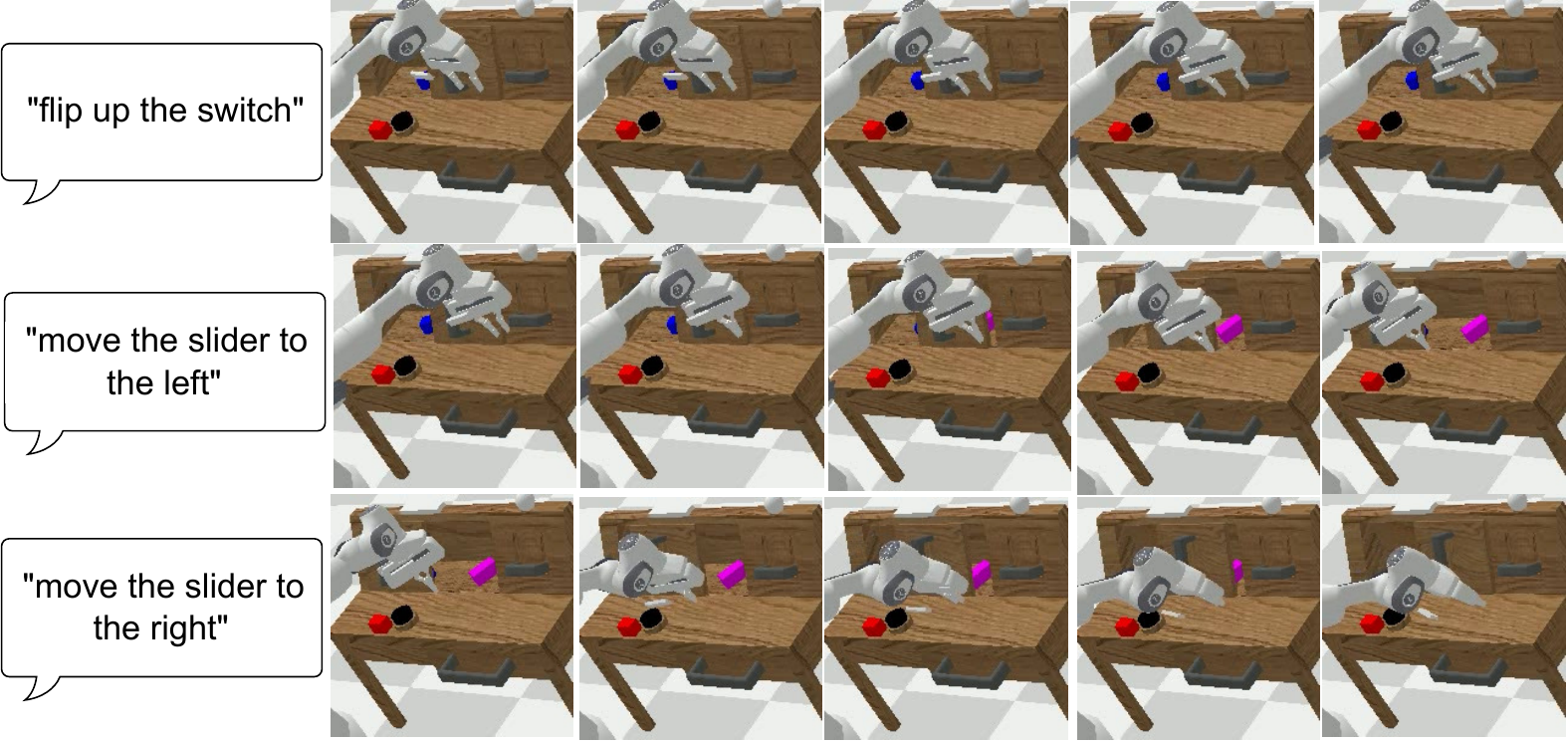}
    \vspace{-0.1cm}
    \caption{Our agent is able to switch between different tasks during inference. We initialize our model with the instruction to flip up the switch on the right side of the table. The frames of the first row depict the agent trying to reach this goal, before the task description changes (second row). The task is then switched to move the slider to the left. After the agent has performed this task, we change the instruction again. The last row depicts the agent trying to move the slider to the right. By the end of the episode (last frame) the slider has returned to its initial position.}
    \label{fig:switching}
    \vspace{-0.6cm}
\end{figure}

\textbf{Single-task performance.} We compare the sample efficiency and success rate of LIMT to the studied baselines. \Cref{fig:individual-tasks} shows the success rate of the different methods on individual tasks over training epochs.
We compute the average success rate of the policies on individual tasks at different evaluation time steps.
LIMT and its variant consistently show better sample efficiency than the baselines.
However, none of the baselines achieve a satisfactory success rate on the \texttt{move\_slider\_right} and \texttt{close\_drawer} tasks.
This illustrates the conflicting objectives problem common to MTRL, since these two tasks might be conflicting with \texttt{move\_slider\_left} and \texttt{open\_drawer} respectively. 
However, one would expect that conditioning the policy on some kind of task representation input would help alleviate this problem.
One explanation is that the embeddings for some tasks such as the \texttt{move\_slider} tasks can be very conflicting and non-discriminatory as shown in \Cref{fig:results}.

To understand whether a model-based approach is beneficial for multi-task learning, we first compare its performance to model-free baselines based on MT-SAC as described in \Cref{sec:baselines}.
The latter do not reach a similar success rate as LIMT under the same training budget. 
To ensure that this comparison is fair, in one variant of this baseline we use the tokenizer from our method as a way to remove the complexity of also learning perception from reward only.
Both MT-SAC:token and MT-SAC:raw lag behind LIMT and its variants.
Furthermore, we aim to validate whether multi-task training is beneficial for individual task performance in MBRL, as was previously shown for model-free methods~\citep{kalashnikov2022scaling}.
Hence, we compare LIMT to the MBRL-ST baseline.
The latter does not benefit from any kind of data or model sharing across tasks.
For most tasks, LIMT's sample efficiency and success rate are substantially higher.
One exception is the \texttt{open\_drawer} task where MBRL-ST shows better sample efficiency than LIMT.
However, such a behavior is to be expected since the single-task policy can more easily excel at learning the one task it specializes in.
In fact, it is a positive sign for our method that this tendency is only seen in one task.
Additionally, we ablate the effect of using the instruction embeddings from the pre-trained language model on LIMT.
Replacing the instruction embeddings with integer task identifiers in both the world model and the actor critic (LIMT:nl) significantly deteriorates the performance in all tasks.
This validates our hypothesis on the importance of using semantically useful task representations for bootstrapping the learning of the dynamics model of similar tasks.
When only removing the language embeddings from the actor-critic (LIMT:nlac), we observe that the resulting agent is competitive with the other baselines, but still lags behind the fully language-informed version of LIMT.

\textbf{Multi-task performance.} In Table~\ref{fig:results}, we compare the multi-task success rate of the different MTRL methods. 
We compute this metric by averaging the success rate of each agent in all the tasks at evaluation time.
LIMT achieves a success rate higher than the model-free baselines by approximately~30\% under the same sample and update budget.
Even when not using language instructions at all in LIMT, the multi-task success rate is still higher than the model-free baselines.
Additionally, we can clearly observe the benefit of using instruction embeddings as task representation to bootstrap the learning of the multi-task world model and actor-critic.
Not using these embeddings in the actor-critic component decreases the success rate by~25\% and not using it at all by~28\%.
These results clearly illustrate the strength of our method for multi-task policy learning.

\textbf{Task Switching.} In \Cref{fig:switching}, we illustrate the emerging capability of LIMT agents to switch to performing new tasks during task execution. 
This behavior can simply be achieved by feeding the policy a different instruction while the agent is performing a given task.
The figure illustrates a successful switching between 3 different tasks without having to reset the agent or the environment. 
We attribute this feature to the relabeling of task data for sharing it across tasks.

\section{Related Work}

\textbf{Multi-task reinforcement learning} is concerned with learning one policy for multiple tasks.
Previous research on this topic highlighted its benefits, such as bootstrapping the learning of more complex tasks~\citep{kalashnikov2022scaling}, but also identified some of its challenges, such as conflicting objectives~\citep{liu2021conflict}.
To address these challenges multiple methods have been proposed.
\citet{rusu2016policydistillation, parisotto2016actormimic} proposed distilling a single multi-task policy from multiple individually trained DQN policies.
\citet{hessel2019multi} demonstrated the first single agent surpassing human performance in the multi-task domain of Atari games~\citep{bellemare2013arcade}. 
The proposed algorithm adapts the contribution of different tasks on the agent's updates in a way that reduces the bias toward specific tasks.
\citet{kalashnikov2022scaling} presented a method for learning multiple manipulation skills using off-policy RL and relabeling shared data across tasks.
\citet{d2020sharing} examined the effect of sharing representation across multiple tasks and demonstrated the benefits of that paradigm on improving the multi-task learning and even its positive effect on individual task performance.
\citet{yang2020multitask} proposed sharing the policy network while separately learning a rerouting mechanism to choose which parts of the network are used for the different tasks.
\citet{tacorl} proposed learning a latent action plan using conditional autoencoders and learning low-level skills conditioned on such low-level plans from an offline dataset, and a high-level policy outputting such high-level actions.

\textbf{Language in robot learning.} Recent work integrated language in robotics for different purposes.
For instance, multiple methods have been proposed for using pre-trained language models or vision-language models~(VLM) as high-level planners for robotics tasks~\citep{ahn2022can,pmlr-v229-rana23a, wake2023chatgpt,singh2023progprompt}.
VLMs have also been used as success detectors~\citep{du2023vision}.
Other work explored the usage of language models as a way to establish communication in multi-agent robotics tasks~\citep{mandi2023roco}.
\citet{karamcheti2022lila} proposed an approach to learn language-informed latent actions as a way to allow humans to influence policy actions.
\citet{driess2023palm} learn an embodied language model integrating sensor measurements and grounding the resulting model with embodiment data.
Multiple efforts have been made to design language-conditioned policies for robotics~\citep{mei2016listen, misra2017mapping, jiang2019language,stepputtis2020language, zitkovich2023rt}.
Similarly, other work has leveraged language embeddings for goal-conditioning of robot policies~\citep{shah2023lm}.

Inspired by previous efforts in model-free MTRL, our work leverages pretrained language models to represent different tasks in multi-task policy learning using a novel model-based approach.

\section{Limitations}
LIMT relies on precomputed language embeddings as a task representation on which the policy is conditioned. As discussed in \Cref{sec:subresults}, these embeddings can be conflicting for tasks with similar textual descriptions but different dynamics. A possible way to alleviate this would be to use our sequence model for finetuning the task embeddings via contrastive representation learning as done in~\citep{mees_what_2022}. Furthermore, model-based trajectory generation only happens during training, while at inference time LIMT samples actions from the policy network without directly accessing the world model. While LIMT's sample efficiency still benefits from its world model, this limits the generalization of our policy when encountering out-of-distribution states. In other works~\citep{diffuser}, trajectory optimization and dynamics learning are more closely coupled.

\section{Conclusion}
\label{sec:conclusion}
We propose a method for learning language-informed multi-task visual world models.
We use a pre-trained language model to embed task instructions into a semantically meaningful latent space and use these embeddings as task representations.
We then use these representations as inputs to the actor, critic and dynamics models as a meaningful task discriminator.
We train all these components using environment interactions in multiple tasks.
Our experiments demonstrate the benefit of model-based training for obtaining multi-task policies, as well as the importance of language conditioning as a way to learn world models and policies in multi-task settings.
Namely, our method substantially outperforms the model-free multi-task baselines demonstrating the benefit of model-based learning for learning multi-task policies.
In addition, we show that multi-task training can lead to a higher success rate than single-task training in the model-based setting and under the same data budget.
Given the high sample complexity of multi-task policy learning, our results provide a promising path towards more efficient learning of such policies based on language-conditioned world models.

\clearpage

\bibliography{bibliography}  %

\clearpage
\appendix

\section{Models and Hyperparameters}

\subsection{Tokenizer}
The tokenizer is a discrete autoencoder that converts image-proprioception pairs into tokenized representations, similar to the one proposed in~\citep{iris}. We downsample the images to a size of $64\times 64$ before feeding them to the network. The tokenizer converts observations into $K = K_x + K_\theta$ tokens of dimension $d_\mathrm{enc}$ using two separate codebooks for images and proprioception, $C_x$ and $C_\theta$ respectively, of vocabulary size $N$. The hyperparameters of the tokenizer are listed in table~\ref{tab:tokenizer}.

\begin{table}[!h]
    \centering
    \caption{Tokenizer hyperparameters}
    \label{tab:tokenizer}
    \begin{tabular}{lc}    
    \toprule
    \textbf{Hyperparameter} &  \textbf{Value}\\
    \midrule
    Vocabulary size ($N$) & 512 \\
    Image tokens ($K_x$) & 16 \\
    Proprioception tokens ($K_\theta$) & 1 \\
    Total tokens ($K$) & 17 \\
    Embedding dimension ($d_{enc}$) & 512 \\
    Frame dimensions ($H,W$) & $64\times 64$ \\
    Encoder/Decoder Layers & 5 \\
    Residual blocks per layer & 2 \\
    Convolution channels & 64 \\
    Self-attention layers at resolution & 8, 16 \\
    Learning rate ($\alpha^\xi$) & 1e-4\\
    \bottomrule
    \end{tabular}
\end{table}

\subsection{World Model}
Our world model is a transformer based on the implementation of IRIS~\citep{iris}. It accepts a sequence of $H(K + 2)$ tensors, including $HK$ tensors of tokenized observations and $2H$ tensors containing the actions and embedded instructions $a_t, l$ for each timestep $0 \leq t < H$. Using an embedding table of size $N \times D$ for the tokens and linear projections for the actions and instructions, we obtain a $H(K+2) \times D$ tensor which is then passed through $L$ GPT2-like~\citep{gpt2} transformer blocks. To predict rewards, episode ends, and observation embeddings, 3 MLP heads follow the transformer blocks.
The world model's hyperparameters are listed in table~\ref{tab:world_model}.

\begin{table}[!h]
    \centering
    \caption{World Model hyperparameters}
    \label{tab:world_model}
    \begin{tabular}{lc}      
    \toprule
    \textbf{Hyperparameter} &  \textbf{Value}\\
    \midrule
    Horizon length ($H$) & 8 \\
    Embedding dimension ($D$) & 512 \\
    Layers ($L$) & 10 \\
    Attention heads & 4 \\
    Embedding dropout & 0.1 \\
    Residual Dropout & 0.1 \\
    Attention Dropout & 0.1 \\
    Weight decay & 1e-4 \\
    Learning rate & 2e-5 \\
    MLP head layers & 3 \\
    Reward loss factor ($\rho$) & 10 \\
    \bottomrule
    \end{tabular}
    
\end{table}

\subsection{Actor Critic}
We implement both the actor and critic networks described in~\ref{sec:algorithm} as $n$-layer MLPs with skip connections of stride 2. At training time, during imagination rollouts, their input is the tokenized representation predicted by the world model, as well as the predicted raw proprioception vectors $\hat{\theta}$. As their dimension is quite small compared to the rest of the inputs (7-dimensional pose) we repeat $\hat{\theta}$ $b$ times before passing it to the MLPs. At inference time, the policy networks receive the tokenized environment observations along with the raw proprioception information as described above. To stabilize training, we maintain a target critic for computing the lambda returns $V_\lambda$, which we periodically update with the weights of our training critic.
Table~\ref{tab:actor_critic} lists the hyperparameters for both networks.

\begin{table}[!h]
    \centering
    \caption{Hyperparameters for both the actor and critic MLPs}
    \label{tab:actor_critic}
    \begin{tabular}{lc}  
    \toprule
    \textbf{Hyperparameter} &  \textbf{Value}\\
    \midrule
    Hidden layers ($n$) & 9 \\
    Neurons per layer & 2048 \\
    Lambda ($\lambda$) & 0.9 \\
    Gamma ($\gamma$) & 0.65 \\
    Learning rate ($\alpha^\phi, \alpha^\psi$) & 4e-5 \\
    Activation & ELU \\
    Critic update interval (training steps) & 200\\
    Proprioception repetitions ($b$) & 10 \\
    Entropy weight ($\eta$) & 1e-4 \\
    \bottomrule
    \end{tabular}
\end{table}

\section{Computational Resources}

We implement LIMT and all baseline models in Python using PyTorch 2.0.1 \citep{pytorch}. The environments of CALVIN use PyBullet for physics simulation. Our experiments run on a GPU cluster managed by the ClearML platform\footnote{clear.ml}, utilizing different GPU models, including NVIDIA A100, NVIDIA V100, and NVIDIA RTX.

\section{Training details}

\subsection{Policy learning}

The critic network $v_\psi$ regresses the value estimates $V_\lambda$ for a horizon length $H$
\begin{equation}
\label{eq:lambda_returns}
    V^{\lambda}_t := r_t + \gamma_t
    \begin{cases}
    (1 - \lambda) v_{\psi}(\hat{s}_{t+1})+ \lambda V^{\lambda}_{t+1}, & t < H \\
    v_{\psi} (\hat{s}_H), & t=H.
    \end{cases}
\end{equation}

The loss function of the actor network with parameter $\phi$ is
\begin{equation}
    L_\phi = - \sum_{t=\tau}^T V^\lambda_t - \eta \mathcal{H}(\pi(a_t | w, l, \theta)),
\end{equation}

where the last term is an entropy objective to encourage exploration.
$w$ is the tokenized observation and $\theta$ the end effector's pose.
The critic's objective is given by
\begin{equation}
\label{critic_loss}
    \min_{\psi} \mathbb{E}_{q_{\theta}, a_{\phi}}
    \left[ \sum_{\tau = t}^{t+H} \norm{ v_{\psi} (s_{\tau}) -  V_{\lambda}(s_{\tau}) }^2
    \right].
\end{equation}

\subsection{Perceptual Loss}
The perceptual loss introduced in \Cref{sec:algorithm} is similar to the one proposed by~\citep{perceptual_johnson}. Given a pre-trained VGG-16 CNN~\citep{vgg}, we select a subset of layers $M$. For each layer $j \in M$, let $\phi_j(x)$ be the activation of the $j$-th layer, in our case a feature map with dimension $C_j \times H_j \times W_j$. We compute the loss between a ground-truth image $x$ and a reconstructed image $x'$ as
\begin{equation}
\label{eq:perceptual_loss}
       L_\mathrm{perceptual}(x,x') = \sum_{j \in M} \frac{1}{C_j H_j W_j} A_j || \phi_j(x) - \phi_j(x') ||^2,
\end{equation}

where $A_j$ are learned affine transformations, implemented as 1x1 convolutions.

\subsection{Online training}
Once the training of our actor-critic begins, we start performing a total of $n_{rollout}$ policy rollouts of $T$ timesteps at each epoch. We perform the same amount of $n_\mathrm{rollout}/6$ rollouts for each task. We append these online episodes to our dataset $\dataset_\mathrm{online}$. We limit the size of $\dataset_\mathrm{online}$ to a maximum of $n_{max}$ episodes. At every epoch we compute the online sampling ratio $p_\mathrm{online}$, described in~\ref{sec:algorithm}
as

\begin{equation}
    p_\mathrm{online} = p_\mathrm{max} \frac{|\dataset_\mathrm{online}|}{n_\mathrm{max}},
\end{equation}

where we set $p_{max}$ as a maximum ratio. When sampling from $D_{online}$, as done in~\citep{iris}, we prioritize later episodes: we divide $D_{online}$ into 4 quarters and we sample 50\% of our episodes from the last quarter, and 25\% from the 3rd quarter. The rest of the episodes is uniformly sampled from the first half.

All training hyperparameters described above, as well as in section~\ref{sec:algorithm} can be found under table~\ref{tab:training_hyperparams}. Our overall training process is further detailed in algorithm~\ref{algo}.

\begin{table}[!h]
    \centering
    \caption{General Training Hyperparameters. The (unnormalized) task weights $w$ correspond to the following tasks in order: \texttt{open\_drawer}, \texttt{close\_drawer},
    \texttt{move\_slider\_left},
    \texttt{move\_slider\_right},
    \texttt{turn\_on\_lightbulb},
    \texttt{turn\_on\_led}.}
    \label{tab:training_hyperparams}
    \begin{tabular}{lc}
    \toprule
    \textbf{Hyperparameter} &  \textbf{Value}\\
    \midrule
    Batch size tokenizer & 64 \\
    Batch size world model & 128 \\
    Batch size actor-critic & 120 \\
    Training steps per epoch & 200 \\
    Optimizer & Adam \\
    Adam $\beta_1$ & 0.9 \\
    Adam $\beta_2$ & 0.999 \\
    Max gradient norm & 10 \\
    Maximum online sampling ratio ($p_{max}$) & 0.5 \\
    Task weights $w$ & [1, 1, 1, 1, 2/3, 2/3] \\
    $N_t$ & 200 \\
    $N_w$ & 100 \\
    $N_f$ & 50 \\
    Relabeling probability ($1 - p$) & 0.15 \\
    Number of rollouts ($n_{rollout}$) & 120 \\
    Timesteps per episode ($T$) & 50 \\
    $n_{max}$ & 5000 \\
    \bottomrule
    \end{tabular}
\end{table}

\begin{algorithm}
    \caption{LIMT}
    \label{algo}
    \begin{algorithmic}[1]
        \State Given:
        \Statex \textbullet pre-trained LLM $\mathcal{L}$
        \Statex \textbullet offline datasets $\dataset_\mathrm{offline}$, $\dataset_\mathrm{filtered}$ of episode trajectories and directives
        \Statex \textbullet initialized discrete autoencoder $A_\xi$, world model $G$, actor $\phi$, critic $\psi$, target critic $\psi_\mathrm{target}$
        \State Encode language directives in $\dataset_\mathrm{offline}$ using $\mathcal{L}$
        \State $\dataset \gets \dataset_\mathrm{offline}$
        \Statex \texttt{//Training tokenizer}
        \State $\dataset_\mathrm{online} \gets \{\}$
        \While {not converged}
            \For {i=1...\texttt{training\_steps}}
            \State sample $B$ states $(x^j,\theta^j)_{j=1}^{B} \sim \dataset$
            \State $\xi \gets \xi - \alpha^\xi \nabla L_A (\theta, x, \theta)$
            \EndFor
            \If {$i > N_t$}
            \Statex \texttt{//Training world model} 
            \For {i=1...\texttt{training\_steps}}
                \State sample $B$ trajectories ${\{(x_t^j, \theta_t^j, a_t^j, r_t^j , \lan^j)_{t=\tau}^{\tau + H}\}}_{j=1}^{B} \sim \dataset$ of $H$ timesteps
                \State Obtain tokens using tokenizer: ${\{(w_t^j, a_t^j, \lan^j)_{t=\tau}^{\tau + H}\}}_{j=1}^B$
                \State predict tokens, rewards and termination $\hat{w}_{t+1}^j$, $\hat{r_t}^j$, $\hat{d_t}^j$
                \Comment{See eq. \eqref{eq:prediction}}
                \State update $G$ using eq. \eqref{eq:loss_world_model}
            \EndFor
            \EndIf 
            \If{$i > N_t + N_w$}
                \State $\dataset \gets \dataset_\mathrm{filtered}$
            \EndIf
            \If{$i > N_t + N_w + N_f$}
            \Statex \texttt{//Training actor critic}
            \For {i=1...\texttt{training\_steps}}
                \State sample $B$ initial states and instruction $(x^j,\theta^j,l^j)_{j=1}^{B (1- p_\mathrm{online})} \sim \dataset$, $(x^j,p^j,l^j)_{j=1}^{B p_\mathrm{online}} \sim \dataset_\mathrm{online}$
                \State relabel language directives $l^j$ with probability $1 - p$
                \State Obtain tokens $\{w_0^j\}^{B}_{j=1}$ using $A_\xi$
                \State Imagine latent trajectories $(w_{t+1}, a_t, r_t, d_t)_{t = 1}^{H-1}$ for each $w^j$ using $G$ and $\phi$
                \State Compute $\lambda$-returns $\{V^\lambda_t\}_{t=1}^H$ using $\psi_\mathrm{target}$
                \State $\phi \gets \phi + \alpha^\phi \nabla \sum_{t=1}^H ( V^\lambda_t +\eta \mathcal{H}(\pi(a_t | w, l, \theta)))$
                \State $\psi \gets \psi - \alpha^\psi \nabla \sum_{t=1}^{H} || V^{\lambda}_t - v_\psi (w_t) || ^2$
            \EndFor
            \State $\psi_\mathrm{target} \gets \psi$
            \Statex  \texttt{//Online rollouts}
                \State perform $n_\mathrm{rollout}$ rollouts of $T$-timestep trajectories $\{\tau^n\}_{n=1}^{n_\mathrm{rollout}}$ using policy $\pi_\phi$
                \State $\dataset_\mathrm{online} \gets \dataset_\mathrm{online} \cup \{\tau^n\}_{n=1}^{n_\mathrm{rollout}}$
            \EndIf
        \EndWhile
    \end{algorithmic}
\end{algorithm}

\subsection{Reward Functions}
The reward functions we specify for our tasks of interest are divided into two classes:

The first one consists of tasks the completion of which can adequately be described by a boolean variable. For example, turning on/off a lightbulb. In these types of tasks, the robot arm has to reach and manipulate an object (e.g a button) to bring about some change in the environment (e.g lightbulb on). A general formula of this type of reward is given in equation \ref{eq:boolean_reward}. 

\begin{equation}
\label{eq:boolean_reward}
    R_{b}(\theta, g, s, l) = 1 - \| (\theta - g) \odot f_s \|_2 + 10\mathbb{I}[\mathrm{success}(s,l)]
\end{equation}

Here, $\theta$ contains is the proprioception vector and $g$ is the desired pose of the end effector to reach an object of interest, such a buttons or a switch. The first term is therefore a distance-based reward. This is employed to guide the agent to the vicinity of the object and avoid relying on sparse rewards, which can slow down the learning process.
We multiply the difference $(\theta - g)$ by a scaling vector $f_s$ to prioritize some proprioception dimensions over others. In our case, end effector orientation is discounted relative to end effector position. The intuition behind this is that, for completing the tasks, it is more important for the EE to reach the specified position of the object of interest, although the orientation also matters to a certain extent. The reason for the latter is that, by constraining the range of possible orientations, we make it easier for the agent to learn suitable actions, while excluding some of the orientations that would make the manipulation of the object significantly harder.

The second term is an indicator variable indicating completion of the task, depending on the environment's state $s \in S$ and the instruction $l$. We multiply it by a constant $\beta_b \in \mathbb{R}$, in our case 10, to amplify the learning signal for successes.

The second class of reward functions are designed for tasks the success of which can best be described as a continuous, time-dependent variable. For example, to define success for the task of opening a drawer, we need information about its (continuous) position in the previous time steps. In these cases, the reward function is defined as follows:

\begin{equation}
\label{eq:continuous_reward}
    R_{c}(p_t, g, s_t, s_{t-1}, l, s_g) = 1 - || (p_t - g) \odot f_s ||_2 + \beta sign(s_g(l) - s_{t-1}) \cdot (s_t - s_{t-1})
\end{equation}

where $\theta_t$ and $g$ are the EE's actual and desired pose, $s_t , s_{t-1}$ are environment states for timesteps $t, t-1$ and $s_g(l)$ is the desired environment state, determined by the instruction $l$. In the example of opening a drawer ,$s_g$ would corrrespond to a fully opened drawer. As above, the first term is a dense distance-based reward. The second term measures progress towards completion of the task during the last timestep and is weighed by a scalar $\beta_c \in \mathbb{R}$ to amplify the learning signal. As the difference $(s_t - s_{t-1})$, can be quite small, we use $\beta_c = 50$.

\begin{table}[!h]
    \centering
    \caption{Reward Hyperparameters. The entries of the scaling factor $f_s$ correspond to the position, orientation and gripper width of the end effector in that order.}
    \label{tab:reward_hyperparams}
    \begin{tabular}{lc}  
    \toprule
    \textbf{Hyperparameter} &  \textbf{Value}\\
    \midrule
    Reward scalar $\beta_b$ & 10 \\
    Reward scalar $\beta_c$ & 50 \\
    Scaling factor $f_s$ & [1, 1, 1, 0.5, 0.5, 0.5, 0] \\
    \bottomrule
    \end{tabular}
\end{table}

\section{Imagined Trajectories}
\begin{figure}[ht]
    \centering
    \begin{tikzpicture}
        \node[anchor= north west] (image1) at (0,0) {\includegraphics[width=\linewidth, trim={6cm 11cm 5cm 8cm}]{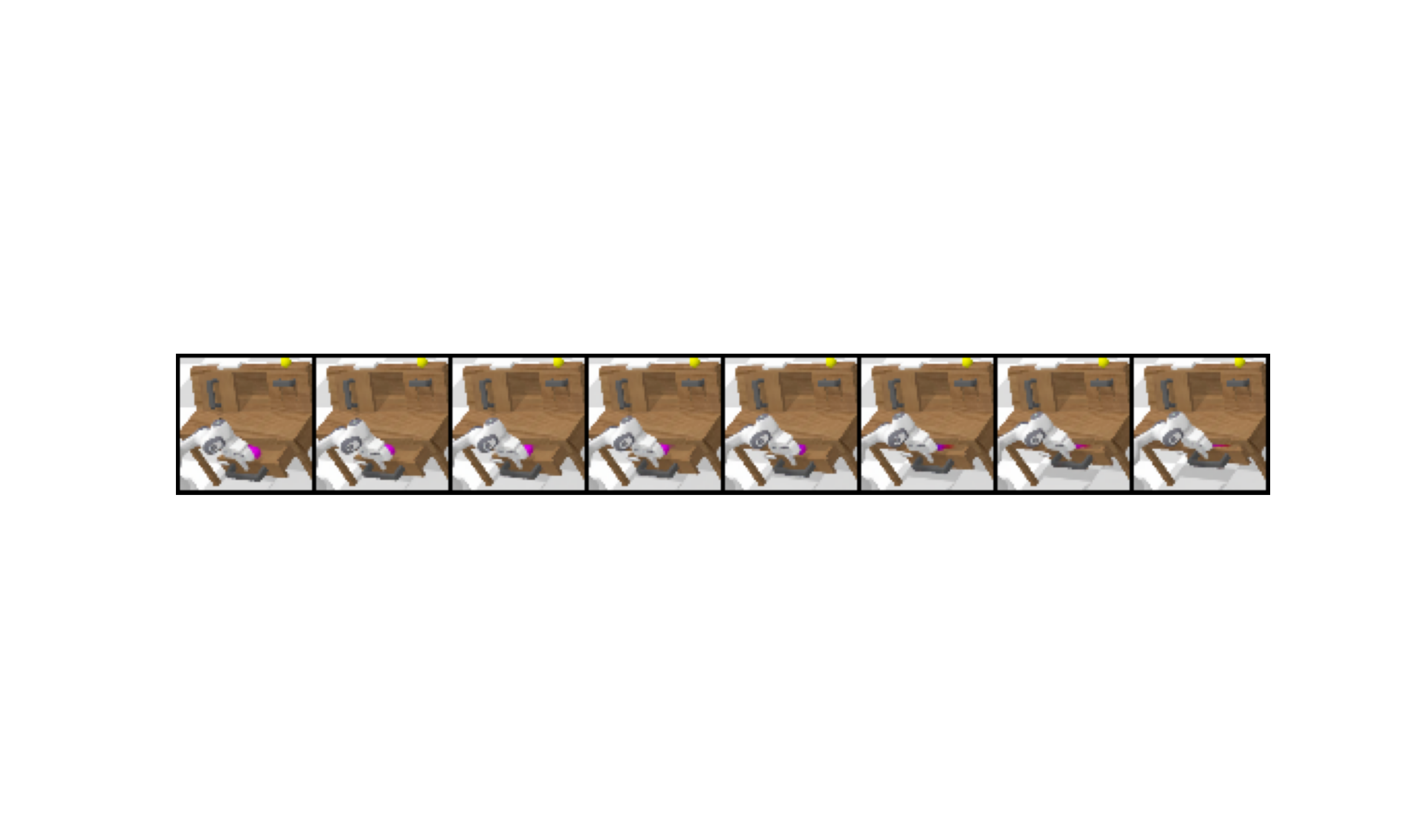}};
        \node[anchor=north west, opacity=0.0] at (0.0, 1.2) {\includegraphics[width=\linewidth, trim={6cm 18cm 5cm 10cm}]{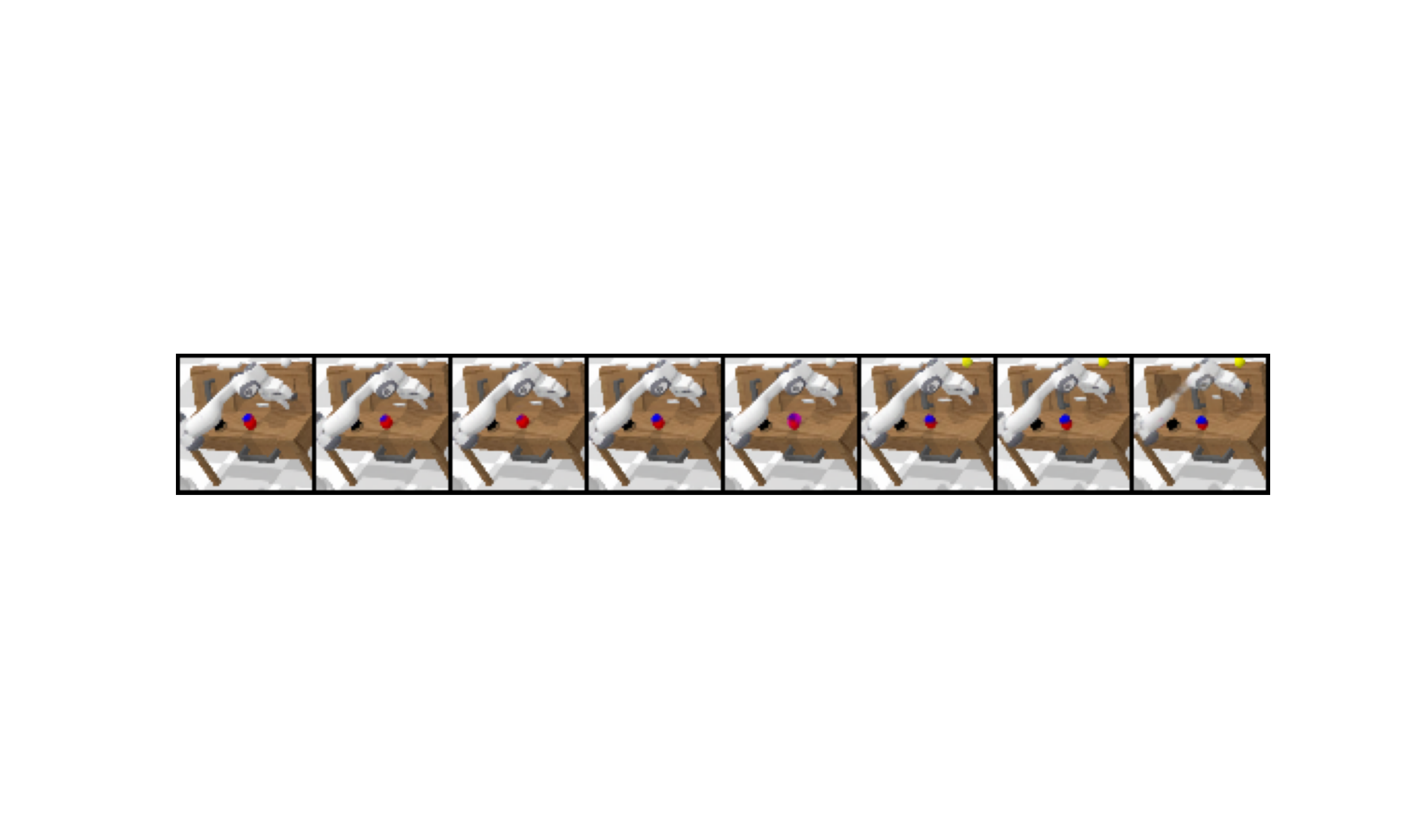}};
    \end{tikzpicture}
    \caption{Reconstructions of imagined trajectories for the tasks \texttt{turn\_on\_lightbulb} (top) and \texttt{close\_drawer} (bottom). The first frame in each row is initialized from the real environment.}
    \label{fig:imagined_trajectories}
\end{figure}

To qualitatively assess our world model, we examine some imagined trajectories generated during policy training. \Cref{fig:imagined_trajectories} illustrates two trajectories of length 8 for the tasks \texttt{turn\_on\_lightbulb} (top) and \texttt{close\_drawer} (bottom). As our agent is trained in latent imagination and not directly on image data, we map the latent outputs to images using the decoder $D$ of our tokenizer for visualization purposes. As can be seen in the figure, our world model successfully captures task-relevant visual features, like the yellow lightbulb lighting up and the drawer transitioning from an opened to a closed position.

\end{document}